\documentclass{article}

\PassOptionsToPackage{numbers, compress}{natbib}

 \usepackage[preprint]{neurips_2025}

\usepackage[utf8]{inputenc} 
\usepackage[T1]{fontenc}    
\usepackage{hyperref}       
\usepackage{url}            
\usepackage{booktabs}       
\usepackage{amsfonts}       
\usepackage{nicefrac}       
\usepackage{microtype}      
\usepackage{xcolor}         
\usepackage{graphicx}
\usepackage{listings}
\usepackage{xcolor}

\title{Latent Graph Learning in Generative Models of Neural Signals}
\workshoptitle{Foundation Models for the Brain and Body}
\author{%
  Nathan X.~Kodama \\
  Case Western Reserve University\\
  Cleveland, OH 44106 \\
  \texttt{nxkodama@case.edu} \\
  \And
  Kenneth A.~Loparo \\
  Case Western Reserve University\\
  Cleveland, OH 44106 \\
  \texttt{kal4@case.edu} \\
}

\begin{document}

\maketitle

\begin{abstract}
  Inferring temporal interaction graphs and higher-order structure from neural signals is a key problem in building generative models for systems neuroscience. 
  Foundation models for large-scale neural data represent shared latent structures of neural signals.
  However, extracting interpretable latent graph representations in foundation models remains challenging and unsolved.  
  Here we explore latent graph learning in generative models of neural signals. 
  By testing against numerical simulations of neural circuits with known ground-truth connectivity, we evaluate several hypotheses for explaining learned model weights. 
  We discover modest alignment between extracted network representations and the underlying directed graphs and strong alignment in the co-input graph representations. 
  These findings motivate paths towards incorporating graph-based geometric constraints in the construction of large-scale foundation models for neural data.
\end{abstract}

\section{Introduction}
Foundation models for neural signals mark a shift from task-specific neural decoders to general-purpose systems trained on heterogeneous neural datasets~\citep{Ye2025, Zhang2024, Wang2024,Azabou2023,Ye2023,Ye2021}. This approach---scaling large pretrained models to capture the complexity, diversity, and population-level structure---expands on learning neural manifolds across tasks, modalities, and animals~\citep{Schneider2023,Gallego2020,Pandarinath2018,Gallego2017}. Recent advances in cell type and brain region identification using \textsc{POYO+}~\citep{Azabou2025} highlight the promise of extracting representations from foundation models with biophysical interpretability.

One promising application of neural foundation models is the identification of network representations from large-scale neural recordings. In particular, Neuroformer has demonstrated the ability to represent functional connectivity, including directionality, via its attention mechanism~\citep{Antoniades2024}. However, the extent to which foundation models learn latent graph representations remains underexplored. 

Several statistical tools have had some success in recovering sign-directed coupling from population activity, including Bayesian state-space methods~\citep{Mishchenko2011}, recurrent neural networks~\citep{Pikovsky2016}, and generalized linear models (GLMs)~\citep{Pillow2008, Truccolo2005, Paninski2004}. Event-timing approaches have further shown what can be recovered from spike interval geometry alone, reconstructing physical connectivity without strong inductive biases~\citep{Casadiego2018}. Our work bridges these areas by treating the transformer's attention as soft adjacency over complete graphs~\citep{Joshi2025} and evaluating whether foundation models internalize network connectivity from fine-to-coarse scales: edge-level reconstruction and higher-order structural alignment.

Here we explore how attention mechanisms in Neuroformer encode connectivity patterns by comparing extracted network representations against GLM baselines and ground truth connectivity from simulated neural circuits. We hypothesize that while foundation models may not perfectly reconstruct edge-level synaptic weights, they may capture higher-order structural regularities that reflect the underlying network organization. 

Our contributions are as follows:
\begin{enumerate}
    \item \textbf{Systematic framework for latent graph extraction}: We explore a principled methodology to extract network representations from the attention mechanisms of Neuroformer and compare them against a well-established baseline (GLMs).
    \item \textbf{Benchmarking against ground truth connectivity}: Using a simulated hub network with known connectivity, we perform a hypothesis-driven evaluation of directed coupling, absolute interaction magnitudes, and higher-order co-input graph structures.
    \item \textbf{Discovery of robust higher-order alignment}: We show that while direct edge-level reconstructions are modestly aligned with ground truth coupling weights, co-input representations exhibit strong alignment, revealing that generative models implicitly encode higher-order latent graph representations.
\end{enumerate}

\section{Methods}
\subsection{Latent Graph Extraction}
\subsubsection{Attention Aggregation from Neuroformer}
Following the methodology established in the original Neuroformer paper~\citep{Antoniades2024}, we analyzed spike-timing data from a recurrent spiking network of $N=300$ leaky integrate-and-fire (LIF) neurons to generate benchmark spike trains with known ground-truth connectivity. Three hub neurons were assigned disproportionately strong outbound synapses, yielding a structured adjacency $A \in \mathbb{R}^{N \times N}$. The resulting dataset contained $\sim 1 \mathrm{M}$ spike events, each represented by its neuron identity and interspike interval. This ground-truth $A$ served as the reference for evaluating model-based connectivity estimates. Details of the hub-network simulation setup are provided in Appendix~\ref{app:hub-net}.

Neuroformer processes spike events $\left\{\left(n_k, t_k\right)\right\}_{k=1}^K$, where $t_k$ is the spike time and $n_k \in\{1, \ldots, N\}$ is the neuron identity. Briefly, each event is mapped to a token corresponding to its neuron identity and inter- or cross-spike interval $\Delta t_k=t_k-t_{k-1}$ and a sequence of tokens of length $T$ is embedded into $Z \in \mathbb{R}^{T \times d}$. 

For layer $\ell$ and head $h$,
$$
Q=Z W_Q^{(\ell, h)}, \quad K=Z W_K^{(\ell, h)}, \quad V=Z W_V^{(\ell, h)}
$$
and the causal attention over tokens is
$$
A^{(\ell, h)}=\mathrm{softmax}\left(\frac{Q K^{\top}}{\sqrt{d_h}}+M\right),
$$
where $M_{k j}=-\infty$ for $j \geq k$ (else 0 ).

During inference, the learned self-attention matrices provide a measure of directed functional influence.  To aggregate to neuron-neuron interactions, define a binary incidence matrix $R \in\{0,1\}^{T \times N}$ with $R_{t, i}=$ 1 iff token $t$ is neuron $i$. The attention mass from presynaptic $j$ to postsynaptic $i$ in that sequence/head/layer is
$$
S^{(\ell, h)}=R^{\top} A^{(\ell, h)} R \in \mathbb{R}^{N \times N}
$$

Across sequences $s$, layers $\ell$, and models $m$, we accumulate total attention and the frequency of causal interactions,
$$
N_a=\sum_{m, s, \ell} S_{(m, s)}^{(\ell, h)}, \quad N_z=\sum_{m, s, \ell}\left(R_{(s)}^{\top} \mathbf{1}_{\mathrm{causal }} R_{(s)}\right),
$$
and report the directed connectivity estimate
$$
C=N_a \oslash N_z,
$$
with $\oslash$ the Hadamard division. Conceptually, $C$ is average attention mass from $j \rightarrow i$ per available causal pairing. A more detailed description of the Neuroformer pipeline is given in Appendix~\ref{app:neuroformer-pipeline}.

\textbf{Baseline GLM Weight Aggregation}: We extract connectivity estimates from generalized linear models by aggregating learned weights that capture sign-directed neuron interactions. This provides a baseline for comparison with the foundation model approach.

For comparison, we implemented a generalized linear model (GLM) framework. Spike trains were discretized into bins of width $\Delta t$, forming a binary matrix $Y_{t, i}$. Presynaptic spike histories were expanded via raisedcosine bases $\left\{\phi_b(\tau)\right\}_{b=1}^B$ :
$$
X_{t, j b}=\sum_{s<t} Y_{s, j} \phi_b(t-s)
$$

For postsynaptic neuron $i$, the conditional intensity was parameterized as:
$$
\lambda_i(t)=\sigma\left(h_i+\sum_{j=1}^N \sum_{b=1}^B W_{i j b} X_{t, j b}\right),
$$
with Bernoulli likelihood over bins. After training, effective filters were reconstructed:
$$
k_{i j}(\tau)=\sum_{b=1}^B W_{i j b} \phi_b(\tau),
$$
and summarized into scalar couplings $J_{i j}$ via the signed peak magnitude of $k_{i j}(\tau)$.
This yielded a weight matrix $J$ that serves as a directed connectivity estimate with explicit sign information. Details of the GLM pipeline are provided in Appendix~\ref{app:glm-pipeline}.

\subsection{Graph Representation Analysis}
\textbf{Ground Truth Comparison}: We evaluate both approaches against the known hub network structure from the original Neuroformer simulation dataset, which has known ground truth connectivity for systematic evaluation. We compare both the directed and absolute magnitudes of graph coupling between predicted and ground truth networks, focusing on the alignment of the attention mechanisms with neural coupling.

\textbf{Co-Input Graph Representations}: We compute the co-input (bibliographic coupling) representations, derived from both model weights and ground truth network coupling, then compare these higher-order graph features that capture shared connectivity patterns.

While edge-level couplings often differ in scale or sign, higher-order structure can be probed via co-input graphs. Given adjacency $A$, the co-input representation is:
$$
B=A A^{\top},
$$
where $B_{i j}$ quantifies similarity in presynaptic inputs to neurons $i$ and $j$.
\begin{itemize}
    \item If both $i$ and $j$ receive excitation from the same source, $B_{i j}>0$.
    \item If both receive inhibition from the same inhibitory source, $B_{i j}>0$.
\end{itemize}
This representation emphasizes shared input structure, providing a more stable latent graph space for comparing model-derived connectivity against ground truth. See Appendix~\ref{app:co-input} for details.

We evaluate latent graph representations using complementary metrics. For the fine-scale edge comparisons, we compute Pearson and Spearman correlations between connectivity matrices; for coarse-cased graph comparisons, we use a graph divergence measure to assess structural differences.

To robustly compare graphs $A$ and $A^{\prime}$, we used a spectral divergence measure based on singular values (to accommodate directed, possibly signed networks). Define:
$$
d\left(A, A^{\prime}\right)=\left\|\sigma(A)-\sigma\left(A^{\prime}\right)\right\|_2,
$$
where $\sigma(\cdot)$ denotes sorted singular values.
To ensure invariance to global scaling and shifts, we center the adjacency matrices by subtracting their mean and normalize by the largest singular value. This produces a stable metric for evaluating differences between latent graphs while discounting global biases in scale or offset. We describe the spectral graph divergence metric in Appendix~\ref{app:spectral-graph-divergence}.

\section{Results}
\subsection{Network Reconstruction}
\begin{figure}[h!]
  \centering
  \includegraphics[width=\textwidth]{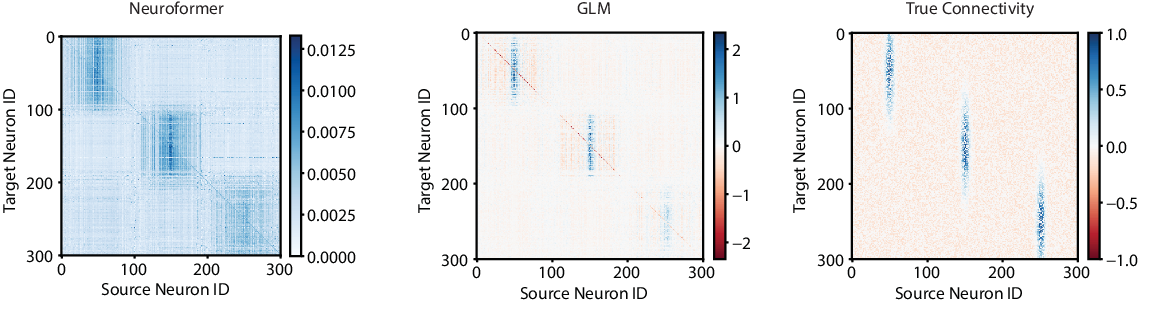}
  \caption{Directed networks from Neuroformer and GLM models compared against ground truth connectivity. (Left): Functional connectivity matrix estimated from aggregated Neuroformer attention weights. (Middle): Connectivity estimated from GLM weight aggregation. (Right) True sign-directed connectivity of the simulated hub network. Color scales indicate interaction strength (blue = excitatory, red = inhibitory}.
  \label{fig:fig-connectivity}
\end{figure}

We compare extracted connectivity matrices from Neuroformer attention aggregation, GLM weight aggregation, and the ground-truth simulated hub network (Figure~\ref{fig:fig-connectivity}). Both approaches recover meaningful directed structure. Neuroformer captures context-dependent couplings, often aligning with temporally structured interactions, though the sign and exact weight magnitudes are less consistent, and the GLM recovers clearer sign-directed interactions. Both exhibit partial alignment with the connectivity of the ground truth network.

\subsection{Comparing Latent Graph Representations}
We assess latent graph representations across three complementary views: (1) True Connectivity ground truth as sign-directed connectivity matrices, (2) Co-input representations that emphasize shared inputs and higher-order structures, and (3) Absolute magnitudes of couplings.

Table~\ref{tab:graph-results} summarizes the comparison between the GLM baseline and Neuroformer, with several trends emerging. Both models achieve modest Pearson and Spearman correlations with the ground truth coupling, with the GLM slightly outperforming Neuroformer (Pearson $r^2=0.262$ vs. $0.110$). Alignment is substantially stronger in the co-input graph representations, particularly for Spearman correlations. Neuroformer achieves the highest performance here ($r^2=0.670$), suggesting that attention excels at capturing higher-order relational structure, even when edge-level estimates are noisy. Finally, both models perform poorly when comparing absolute magnitudes of edge-weights, with low correlations and large graph spectral distances.

These findings offer a key insight: while exact edge weights are difficult to recover, both models---especially, Neuroformer---capture higher-order structural regularities in the network. This suggests that generative models intrinsically internalize the co-input graph space, providing a robust representation of collective connectivity patterns.

\begin{table}[h]
\centering
\caption{Comparison of network reconstruction quality between GLM and Neuroformer models. 
Metrics are reported for true connectivity, co-input representations, and absolute magnitude coupling.}
\label{tab:graph-results}
\begin{tabular}{lccc|ccc}
\toprule
 & \multicolumn{3}{c}{GLM} & \multicolumn{3}{c}{Neuroformer} \\
\cmidrule(lr){2-4} \cmidrule(lr){5-7}
 & True Conn. & Co-Input & Abs. Mag. & True Conn. & Co-Input & Abs. Mag. \\
\midrule
Pearson $r^2$ & 0.262 & \textbf{0.481} & 0.200 & 0.110 & \textbf{0.536} & 0.040 \\
Spearman $r^2$ & 0.164 & \textbf{0.636} & 0.041 & 0.132 & \textbf{0.670} & $-0.007$ \\
Graph Spectral Dist. & 0.767 & \textbf{0.318} & 1.742 & 1.257 & \textbf{0.476} & 2.075 \\
\bottomrule
\end{tabular}
\end{table}

\section{Discussion}
Our findings show that both Neuroformer and GLM-based approaches can recover aspects of latent network structure, though with different strengths. GLMs offer clear sign-directed couplings, while Neuroformer’s attention patterns capture context-dependent dependencies. Direct edge-level alignment with ground truth was modest in both cases, reflecting the intrinsic difficulty of network reconstruction.

The strongest result arises in the co-input representation space: Neuroformer’s attention consistently aligned with higher-order connectivity patterns, even when edge magnitudes were noisy. This suggests that generative models do not simply approximate direct synaptic weights but instead internalize collective input–output regularities.

Taken together, these results highlight complementary advantages: GLMs provide interpretable directed weights, while transformer-based models reveal higher-order latent graphs that better explain observed dependencies. Scaling these approaches will require methods to incorporate graph constraints efficiently into large models and to evaluate across diverse neural modalities.

\section{Conclusion}
We demonstrate that foundation models of neural data contain extractable latent graph representations with meaningful biophysical interpretations. Through systematic comparison with GLM baselines on a hub-network simulation, we show three main contributions:
\begin{enumerate}
    \item A principled framework for extracting directed connectivity from generative models of neural signals.
    \item Benchmarks against ground-truth connectivity, highlighting trade-offs between direct coupling estimates and higher-order graph features.
    \item Discovery of strong alignment in co-input graph spaces, revealing that generative models implicitly encode shared-input structures.
\end{enumerate}
These findings suggest that foundation models can serve both as predictive tools and as instruments for structure discovery. As models scale to larger datasets, the ability to extract latent representations will be critical for bridging predictive performance with biophysical interpretability.

\newpage
\bibliographystyle{unsrt}
\bibliography{references}{}

\newpage
\appendix
\section{Hub-Network Simulation Setup}\label{app:hub-net}
The ground-truth synaptic matrix included three "hub" neurons with disproportionately strong outgoing connections to many other neurons. These hubs acted as key drivers of network activity. A recurrent spiking network of $N=300$ leaky integrate-and-fire (LIF) neurons. Each neuron's membrane potential $V_i(t)$ followed the stochastic differential equation
$$
\frac{d V_i}{d t}=\frac{I-V_i}{\tau}+\sigma \sqrt{\frac{2}{\tau}} \xi_i(t)
$$
where $\tau=10 \mathrm{~ms}$ is the membrane time constant, $\xi_i(t)$ is a Gaussian random variable with zero mean and unit variance,  $\sigma=0.5$ scales the stochastic drive 0. The resulting spike train dataset contained about 1 million tokens (spike events, represented as neuron IDs with time intervals). Three hub neurons with strong outbound connectivity, $\tau=10 \mathrm{~ms}, \sigma=0.5,1 \mathrm{M}$ tokens dataset. Used to validate Neuroformer's ability to infer directed connectivity via attention.

\section{Description of the Neuroformer Pipeline}\label{app:neuroformer-pipeline}

\subsection{Tokenization and Embedding}
We observe a population of $N$ neurons with spike events $\left\{\left(n_k, t_k\right)\right\}_{k=1}^K$, where $t_k$ is the spike time and $n_k \in\{1, \ldots, N\}$ is the neuron identity. Each event is mapped to a token corresponding to its neuron identity and inter- or cross-spike interval $\Delta t_k=t_k-t_{k-1}$. Thus, an input sequence of tokens of length $T$ is represented as $x= \left\{\left(n_k, \Delta t_k\right)\right\}_{k=1}^T$. 

Each token is transformed by a learned embedding $\mathbf e_{n_k} \in \mathbb{R}^d$ and added together with a Fourier temporal embedding $\mathbf u\left(\Delta t_k\right)$. The token embeddings are then stacked:
$$
Z^{(0)}=\left[\begin{array}{c}
\mathbf z_1 \\
\vdots \\
\mathbf z_T
\end{array}\right] \in \mathbb{R}^{T \times d}, \quad \mathbf z_k=\mathbf e_{n_k}+\mathbf u\left(\Delta t_k\right)
$$

\subsection{Causal Transformer Blocks}
In Neuroformer, $L$ causal transformer blocks are stacked. For block $\ell=1, \ldots, L$ with $H$ heads and per-head width $d_h\left(H d_h=d\right)$, the query, keys, and values are
$$
Q^{(\ell, h)}=Z^{(\ell-1)} W_Q^{(\ell, h)}, \quad K^{(\ell, h)}=Z^{(\ell-1)} W_K^{(\ell, h)}, \quad V^{(\ell, h)}=Z^{(\ell-1)} W_V^{(\ell, h)}
$$
and the attention matrix and output are
$$
A^{(\ell, h)}=\mathrm{softmax}\left(\frac{Q^{(\ell, h)} K^{(\ell, h) \top}}{\sqrt{d_h}}+M\right), \quad O^{(\ell, h)}=A^{(\ell, h)} V^{(\ell, h)},
$$
where $M$ is the attention mask, which enforces causality, by zeroing out disallowed key positions.

Outputs are concatenated across heads and projected
$$
O^{(\ell)}=\left[O^{(\ell, 1)}\|\cdots\| O^{(\ell, H)}\right] W_O^{(\ell)},
$$
and then residual, layer normalization, MLP, and second residual are applied to obtain $H^{(\ell)}$.

\subsection{Output Heads}
Given $H^{\mathrm{out }}=H^{(L)}$, Neuroformer predicts the next neuron identity and interspike interval with softmax heads:
$$
p\left(n_{k+1} \mid x_{\leq k}\right)=\mathrm{softmax}\left(H_k^{\mathrm{out }} W_n\right), \quad p\left(\Delta t_{k+1} \mid x_{\leq k}\right)=\mathrm{softmax}\left(H_k^{\mathrm{out }} W_t\right)
$$
Training minimizes the weighted sum of cross-entropy losses across both outputs.

\subsection{Attention-Based Connectivity Extraction}
During inference, the learned self-attention matrices provide a measure of directed functional influence. Let $A^{(\ell, h)} \in \mathbb{R}^{T \times T}$ be the causal attention matrix for a sequence.
Define an incidence matrix $R \in\{0,1\}^{T \times N}$ with $R_{t, i}=1$ iff token $t$ is neuron $i$. The sequence-level neuron attention flow for layer $\ell$, head $h$ is
$$
S^{(\ell, h)}=R^{\top} A^{(\ell, h)} R \in \mathbb{R}^{N \times N} .
$$

For each neuron pair, we aggregate over sequences $s$, layers $\ell$,  and models $m$ :
$$
N_a=\sum_{m, s, \ell} S_{(m, s)}^{(\ell, h)}, \quad N_z=\sum_{m, s, \ell}\left(R_{(s)}^{\top} \mathbf{1}_{\mathrm {causal }} R_{(s)}\right),
$$
where $\mathbf{1}_{\mathrm {causal }} \in\{0,1\}^{T \times T}$ has ones for allowed key positions in the past, $N_a$ is the average attention weight, and $N_z$ is the frequency of neuron $j$ being attended when predicting neuron $i$. The final directed latent connectivity is
$$
C=N_a \oslash N_z \in \mathbb{R}^{N \times N}
$$

\subsection{Training Details}
Training was carried out end-to-end using the Adam optimizer by minimizing the sum of cross-entropy losses for neuron identity prediction and interspike interval prediction. We trained $4$ independent models and averaged their extracted attention maps for robust connectivity estimation.

The training setup includes the following configuration:
\begin{itemize}
    \item \textbf{Data split}: Spike trains are split chronologically into training and testing subsets: an $80 / 20$ split was used for all datasets.
    \item \textbf{Mini-batch size}: Training is carried out with mini-batches of 64 sequences.
    \item \textbf{Learning rate}: fixed at $1 \times 10^{-3}$.
    \item \textbf{Regularization}: Dropout layers are applied at embeddings, attention, and residual connections (default 0.1).
\end{itemize}

\section{ Description of the GLM Pipeline}\label{app:glm-pipeline}
\subsection{Spike Train Representation}
We record spike times from $N$ neurons over a simulation of duration $T$ (ms). Let
$$
\left\{t_k^{(i)}\right\}_{k=1}^{K_i}, \quad i=1, \ldots, N
$$
denote the set of spike times (in seconds) for neuron $i$. We bin spike trains with bin size $\Delta t = 1$ ms, forming a binary spike matrix
$$
Y_{t, i} \in\{0,1\}, \quad t=1, \ldots, T_{\mathrm{bins }}, \quad i=1, \ldots, N,
$$
where $Y_{t, i}=1$ if neuron $i$ spikes in bin $t$, else 0.

\subsection{Basis Expansion of Spike History}
Each presynaptic spike train is convolved with a set of raised-cosine basis functions $\left\{\phi_b(\tau)\right\}_{b=1}^B$, defined on a time window of length $L$ (samples). For each neuron $j$ and basis $b$ :
$$
X_{t, jb}=\sum_{s<t} Y_{s, j} \phi_b(t-s)
$$
where $Y_{s, j}$ is the binary spike indicator of neuron $j$ at time bin $s$ and $\phi_b(\cdot)$ encodes temporal filters (learned as a linear combination through $W_{i j b}$ ).

The resulting design matrix has shape
$$
X \in \mathbb{R}^{T_{\mathrm {bins }} \times(N \cdot B)}
$$

\subsection{Likelihood and Loss}
For each neuron $i$, spikes are modeled as conditionally independent Bernoulli events per bin. The log-likelihood is
$$
\mathcal{L}=\sum_{t=1}^{T_{\mathrm {bins }}} \sum_{i=1}^N Y_{t, i} \left[ \log \lambda_i(t)- \lambda_i(t) \Delta t\right]
$$
which corresponds to the discrete-time hazard loss that was implemented. Training minimizes $-\mathcal{L}$ by gradient descent (via TensorFlow/Keras).

\subsection{Weight Extraction and Filter Reconstruction}
After training, the dense layer stores weights
$$
W \in \mathbb{R}^{(N \cdot B) \times N}
$$
We reshape into
$$
W_{i j b}, \quad i=1, \ldots, N, j=1, \ldots, N, b=1, \ldots, B
$$
interpretable as coupling strength from presynaptic neuron $j$ to postsynaptic neuron $i$ on basis component $b$. The time-domain coupling filters are then reconstructed as
$$
k_{i j}(\tau)=\sum_{b=1}^B W_{i j b} \phi_b(\tau), \quad \tau=1, \ldots, L
$$
where $k_{i j}(\tau)$ is the effective post-synaptic potential (PSP)-like filter from $j$ to $i$.

To extract a single scalar coupling strength per pair $(i, j)$, one common choice is the signed peak magnitude of the reconstructed filter:
\[
J_{ij} =
\left\{
\begin{array}{ll}
\displaystyle \max_{\tau}\, k_{ij}(\tau), &
\textrm{if } \displaystyle \max_{\tau} |k_{ij}(\tau)| = \max_{\tau} k_{ij}(\tau), \\[6pt]
\displaystyle \min_{\tau}\, k_{ij}(\tau), &
\textrm{otherwise.}
\end{array}
\right.
\]
This ensures that the effective weight reflects the strongest excitatory (positive) or inhibitory (negative) effect across the filter.

\subsection{Training Details}
The model parameters (weights $W_{i j b}$ and biases $h_i$) were estimated by minimizing the negative log-likelihood using gradient descent in TensorFlow. Training was carried out with the following configuration:
\begin{itemize}
    \item Data split: Spike train data were split chronologically into $80 \%$ training and $20 \%$ validation.
    \item Batching: Training used the full sequence per gradient step (i.e., no minibatching).
    \item Epochs: Training ran for up to 20,000 epochs, with early stopping.
    \item Optimizer: Gradient updates were applied using the default TensorFlow optimizer (Adam or SGD depending on configuration).
    \item Learning rate schedule:
    \begin{itemize}
        \item The learning rate was initialized at its TensorFlow default (typically $10^{-3}$ ).
        \item If the validation loss failed to improve over 20 epochs, the learning rate was reduced by a factor of 0.1.
        \item A minimum learning rate of $10^{-5}$ was enforced.
    \end{itemize}
    \item Early stopping:
    \begin{itemize}
        \item Training was stopped if the validation loss failed to improve over $100$ consecutive epochs. An improvement was only recognized if the decrease in validation loss exceeded $\Delta=10^{-6}$.
        \item Best weights: At the end of training, the model parameters were restored to the best checkpoint observed on the validation set.
    \end{itemize}
\end{itemize}

\section{Co-Input Graph Representation}\label{app:co-input}
The co-input graph representation, also known as the bibliographic coupling graph or the dual of co-citation, links neurons that share inputs. If the presynaptic indices are in the columns of the connectivity matrix $A$, the co-input graph is defined as
$$
B=A A^{\top}
$$
$B_{i j}$ measures how much neurons $i$ and $j$ share the same presynaptic sources. The co-input graph is symmetric and, under Dale's principle, it is non-negative because columns have constant sign.

The co-input graph captures higher-order structural relationships. In spiking data, prediction works well if two neurons have correlated input streams (common drive). Neuroscience interpretation:
\begin{itemize}
    \item If two postsynaptic neurons receive excitation from the same excitatory neuron, their entry is positive.
    \item If they both receive inhibition from the same inhibitory neuron, also positive.

\end{itemize}
This captures correlation of input patterns across neurons. It answers: "Which neurons are co-recipients of similar presynaptic sources?" Attention in transformer modes captures which past neurons are predictive of a current spike. Attention-derived $\hat{C}$ will often reflect statistical dependency from co-inputs, not the signed synapse directly. 

\section{Spectral Graph Divergence Measure}\label{app:spectral-graph-divergence}
We developed a spectral divergence measure between graphs through their spectral norm distance:
$$
d_\sigma(A, B)=\|\lambda(A)-\lambda(B)\|_2
$$
where $\lambda(\cdot)$ are sorted eigenvalues. There are several challenges in comparing sign-directed neuronal networks:
\begin{itemize}
    \item Adjacency $A$ is not symmetric, so eigenvalues can be complex.
    \item Comparing eigenvalues directly $(\lambda(A)) \rightarrow$ ordering is ambiguous, distance can be weird (complex plane).
    \item Laplacian definitions ($L=D-A$) are not guaranteed to be symmetric or positive semidefinite, so the "nice" theory of normalized Laplacian doesn't carry over.
\end{itemize}

Instead of eigenvalues, we compute singular values $\sigma(A)$, which are always real and nonnegative:
$$
d(A, B)=\|\sigma(A)-\sigma(B)\|_2
$$
This is the most stable and meaningful generalization.

As is, the spectral graph divergence measure is not invariant to global shifts or rescalings of the adjacency matrix. If you subtract the min, or take abs, the distances change - even though the relative structure is the same. Global shifts are defined as
\begin{itemize}
    \item Shift (additive): $A^{\prime}=A+c \cdot \mathbf{1 1}^{\top}$. Every entry gets a constant $c$.
    \item Scale (multiplicative): $A^{\prime}=\alpha A$. Every entry is multiplied by the same factor. For structural similarity, we impose invariance to both.
\end{itemize}

\textbf{Making metrics shift-invariant} by centering each matrix, i.e. subtract means:
$$
\tilde{A}=A-\frac{1}{n^2} \sum_{i j} A_{i j}
$$
Then compute divergences on $\tilde{A}, \tilde{B}$.
$\rightarrow$ Removes global bias offset.

\textbf{Making metrics scale-invariant} by normalizing by the largest singular value:

$$
\hat{A}=\frac{A}{\sigma_{\max }(A)}.
$$

Our code implementation of the spectral graph divergence measure is given in Python.
\lstset{
  language=Python,
  basicstyle=\ttfamily\footnotesize,
  keywordstyle=\color{blue},
  commentstyle=\color{gray},
  stringstyle=\color{orange},
  showstringspaces=false,
  breaklines=true,
  frame=single
}

\begin{lstlisting}[language=Python, caption={Directed spectral distance with shift and scale invariance}]
def directed_spectral_distance_invariant(A, B, k=None):
    # Center (shift invariance)
    A = A - A.mean()
    B = B - B.mean()

    # Normalize by largest singular value (scale invariance)
    sA = np.sort(la.svdvals(A))[::-1]
    sB = np.sort(la.svdvals(B))[::-1]

    sA /= sA[0] if sA[0] > 0 else 1
    sB /= sB[0] if sB[0] > 0 else 1

    n = min(len(sA), len(sB))
    if k is not None:
        n = min(n, k)
    return la.norm(sA[:n] - sB[:n])
\end{lstlisting}

\end{document}